# Teleoperation System for Service Robots using Virtual Reality Headset and Pose Estimation

Eduardo Fernandes
Lab. Automação e Robótica
Centro Algoritmi
University of Minho
Guimarães, Portugal
a84063@alunos.uminho.pt

Tiago Ribeiro
Lab. Automação e Robótica
Centro Algoritmi
University of Minho
Guimarães, Portugal
id9402@alunos.uminho.pt

António Ribeiro
Lab. Automação e Robótica
Centro Algoritmi
University of Minho
Guimarães, Portugal
afaribeiro00@gmail.com

Carolina Lopes
Lab. Automação e Robótica
Centro Algoritmi
University of Minho
Guimarães, Portugal
carolina.coelho.lopes.28@gmail.com

Gil Lopes
LIACC
ISEP-IPP
Porto
Portugal
agl@isep.ipp.pt

Fernando Ribeiro
Lab. Automação e Robótica
Centro Algoritmi
University of Minho
Guimarães, Portugal
fernando@dei.uminho.pt

**Abstract:** This paper presents a teleoperation system for service robots that integrates human pose estimation with virtual reality (VR) to enable intuitive remote control. By using the Mediapipe framework for 2D pose detection and extending it with RGB-D data from the Intel RealSense D455 camera, a comprehensive 3D human pose estimation is achieved. This data is then mapped to the robot's kinematics, allowing the operator to control the robot's movements naturally. A VR interface was developed to provide an immersive perspective, enhancing the operator's situational awareness. Real-world experiments were conducted using the CHARMIE service robot to validate the system's effectiveness in tasks involving object manipulation. While the current system successfully demonstrates the potential of VR-based teleoperation, challenges remain in accurately capturing certain joint angles (e.g., wrist yaw) and ensuring precise mapping between human and robot movements. Future work includes the integration of inertial measurement units (IMUs) for improved rotational tracking and the implementation of more advanced control algorithms. Overall, this research contributes to the development of more intuitive and effective teleoperation systems for service robots, facilitating their deployment in complex and dynamic environments. This work further introduces a novel approach to teleoperation by combining RGB-D sensing with an innovative algorithm for real-time joint angle computation using Mediapipe, enhancing the precision of human-robot motion mapping and enabling immersive control through a VR interface. These contributions represent significant advancements in the field of human-robot interaction, particularly for service robot applications.

## 1. Introduction

Service robots are increasingly being integrated into diverse applications, including assistance, entertainment, and education. A crucial aspect of these robots is their ability to be controlled remotely, either through user interfaces or by interpreting human gestures and movements. Given the rapid evolution of the service robotics market, the term "service robot" still lacks a universally accepted definition [1]. Nonetheless, the International Federation of Robotics (IFR) defines service robots as semi- or fully-autonomous systems designed to provide useful services that enhance human welfare, explicitly excluding manufacturing operations [2].

As service robots take on increasingly complex and sometimes high-risk tasks, there is a growing need for them to operate under human supervision through teleoperation rather than being fully autonomous. Teleoperation, often referred to as "remote operation," enables a human operator to interact with the robot from a distance—where both "work" and "distance" can vary considerably depending on the context [3]. It can be understood as an extension of human perception and manipulation capabilities, enabling control via communication channels that connect human operators to the robot's sensors and actuators. Typically, teleoperation involves direct and continuous human control of the teleoperated system [4].

Depending on the level of human control, teleoperation can be classified into three categories: direct, coordinated, and task-based [3]. In direct teleoperation, the operator directly controls the actuators, receiving real-time feedback—an example being a remote-controlled car. In coordinated teleoperation, the operator still controls the actuators but benefits from assistance through a remote-control loop, such as setting the speed of a remotely operated machine. In task-based teleoperation, the teleoperated system can perform tasks autonomously while the operator provides high-level commands via a human-machine interface (HMI); for example, a robot performing a semi-autonomous task with operator oversight and adjustment.

One of the enabling technologies for effective teleoperation is pose estimation. Pose estimation refers to the process of determining the configuration of a person's body in an image, including the identification of key points such as facial landmarks, hands, and body joints. Complementing this, Virtual Reality (VR) technology enhances teleoperation by immersing the user in a simulated environment that feels realistic and interactive. Rather than interacting through a traditional computer screen, users are immersed in the interface, and their physical movements are captured, facilitating natural human-robot interaction [5]. VR typically involves a combination of computer-generated real-time animations, motion-sensing devices like wired gloves, and head-mounted displays (HMDs) that provide a stereoscopic visual output, offering users a rich and immersive experience [6] [7].

This study is part of the CHARMIE project [8] (Collaborative Home/Healthcare Assistant Robot by Minho Industrial Electronics), developed at the Laboratory of Automation and Robotics at the University of Minho. The primary objective is to develop an intuitive interface that allows human operators to effectively control the CHARMIE robot through teleoperation. To achieve this, all sensory data—from cameras to sensors—is transmitted and displayed to the user in VR. The robot is then controlled through a combination of pose estimation and computer vision systems that interpret human body joint positions and translate them into robotic actions. This enables the robot to perform various tasks, including object manipulation, thereby bridging the gap between human intention and robotic execution.

## 2. State of the Art

The integration of virtual reality (VR) into the teleoperation of service robots has significantly enhanced remote human–robot interaction. This advancement facilitates intuitive control, immersive feedback, and improved task performance across various applications, including manufacturing, hazardous environment operations, and assistive services. Below is a summary of four pivotal studies contributing to this field:

### 2.1 MIT CSAIL's Homunculus Model for VR Teleoperation

Researchers at MIT's Computer Science and Artificial Intelligence Laboratory (CSAIL) developed a VR system that immerses operators within a robot's control interface, enabling intuitive manipulation of robotic limbs using VR headsets like the Oculus Rift. This system demonstrated superior performance in object manipulation tasks compared to traditional control methods, highlighting the potential of VR to enhance teleoperation efficiency and user experience [9].

### 2.2 Five-Dimensional Human–Robot VR Teleoperation System

A study published in *Robotics* introduced a comprehensive VR-based teleoperation framework that addresses control, visualization, interaction, usability, and infrastructure. This system allows operators to perform complex tasks remotely with minimal prior experience, emphasizing adaptability and immersive interaction. The integration of these five dimensions aims to overcome limitations in existing teleoperation systems by providing a more holistic and user-friendly approach [10].

### 2.3 Virtual Reality Interface for Industrial Robot Manipulators

Researchers explored the application of commercially available VR interfaces for teleoperating industrial robot manipulators. The study identified challenges in applying standard VR control practices to industrial platforms, such as limited access to torque and velocity control. To address these issues, the authors proposed a simplified filtering approach to process command signals, enabling effective teleoperation of industrial robots in dexterous manipulation tasks [11].

### 2.4 Immersive Virtual Reality Bimanual Telerobotic System with Haptic Feedback

A recent study presented a bimanual telerobotic system that combines immersive VR with haptic feedback to enhance operator experience. The system features dual-arm and dual-dexterous hands, providing visual and force feedback to the operator. Experimental results indicated that haptic feedback improves physical perception and task performance, compensating for visual perception deficiencies and reducing operator workload [12].

These studies collectively demonstrate the significant strides made in VR-based teleoperation of service robots, highlighting the importance of immersive interfaces, comprehensive system design, and multimodal feedback in enhancing remote robotic operations.

## 3. Algorithm Architecture: Pose Estimation Model

The estimation of human pose in three dimensions from images or video has attracted significant attention in recent years, largely due to the proliferation of new applications such as human-robot interaction, virtual reality gaming, and sports performance monitoring. Despite substantial progress and numerous studies in the field, the problem remains an active area of research, with opportunities for further improvement and refinement.

In this work, the proposed pose estimation model leverages Google's open-source software MediaPipe [14], which offers a robust and integrated solution for detecting hands, face, and full-body pose concurrently. MediaPipe employs a holistic model pipeline that combines separate models for body pose, face, and hands into a unified framework, thereby enabling simultaneous detection of multiple key points with high efficiency.

The mapping of human body posture to robot joint motion is performed using a systematic approach that involves several steps. First, the 3D positions of human key points (e.g. shoulders, elbows, wrists) are estimated using Mediapipe with depth data integration. Next, each human joint angle (e.g. shoulder pitch) is calculated using vector mathematics (e.g. scalar product or cross product). These calculated human joint angles are then mapped to the robot's corresponding actuators using a direct proportional mapping or a linear scaling function. For example, a detected shoulder pitch angle is scaled to match the physical limits of the robot's shoulder joint and then sent as a command to the respective servo motor. This ensures that the robot's joint angles reflect the human pose in real time.

### 3.1 From 2D to 3D

MediaPipe predicts a landmark model consisting of 33 key points, each with three degrees of freedom: x, y, and visibility. However, this model is inherently limited to two-dimensional predictions, which is insufficient for accurately capturing all operator movements in three-dimensional space. To enhance this model, it is necessary to add a third degree of freedom—depth—to each key point, thereby enabling a more precise and complete representation of the operator's pose.

To address this issue, an RGB-D camera is introduced, which allows each key point to acquire depth (z coordinate) information. A comparative study was conducted using two RGB-D cameras: the Intel® RealSense™ D455 and the Microsoft Kinect XBOX 360, evaluating their respective performance in depth mapping. Figure 1 illustrates the differences between the two systems: the Intel depth map employs a color-coded representation to denote distance, while the Kinect depth map uses grayscale shading, where darker colours indicate objects further from the camera.

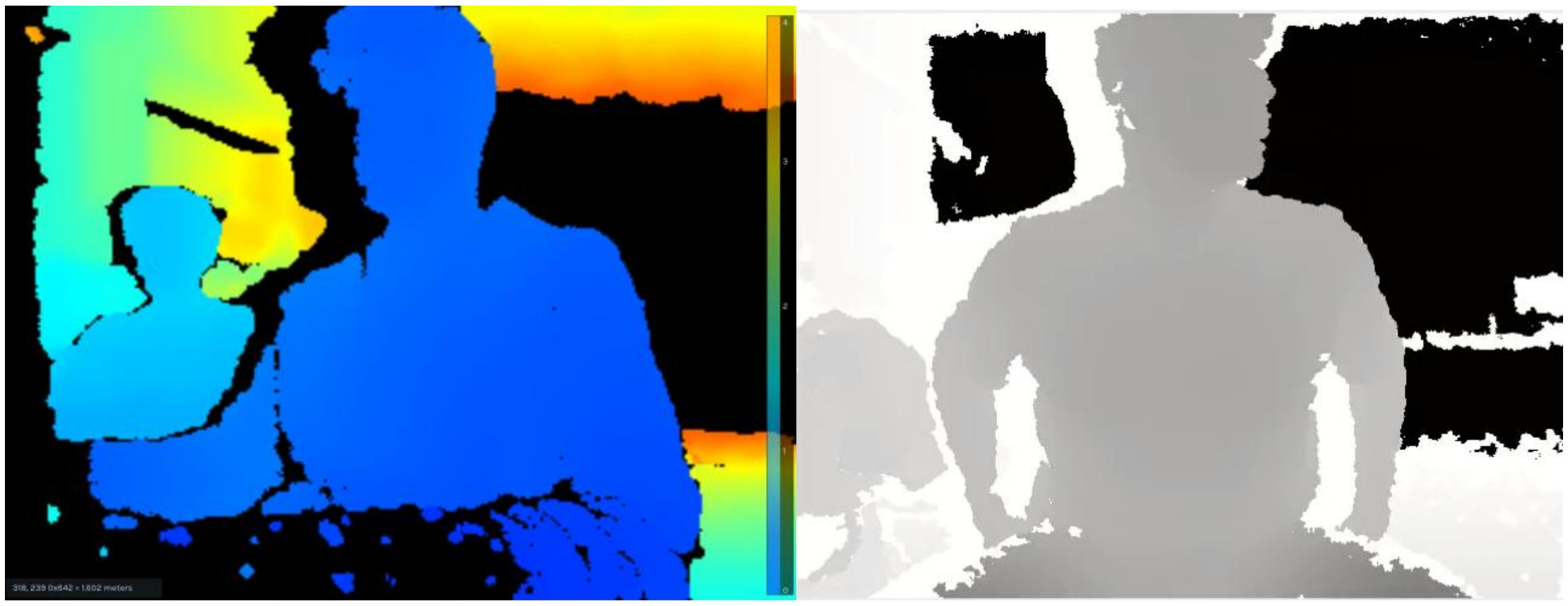

*Figure 1 - Intel® RealSense™ D455 (left) and Microsoft Kinect XBOX 360 (right) depth maps*

The Intel RealSense technology features a built-in function called *get_distance()*, which directly returns the depth value in meters, offering a straightforward and accurate solution. In contrast, the Kinect system requires a more complex process involving the conversion of grayscale values to depth, which entails measuring distances at multiple points to build a calibration curve and its corresponding function, as shown in Figure 2.

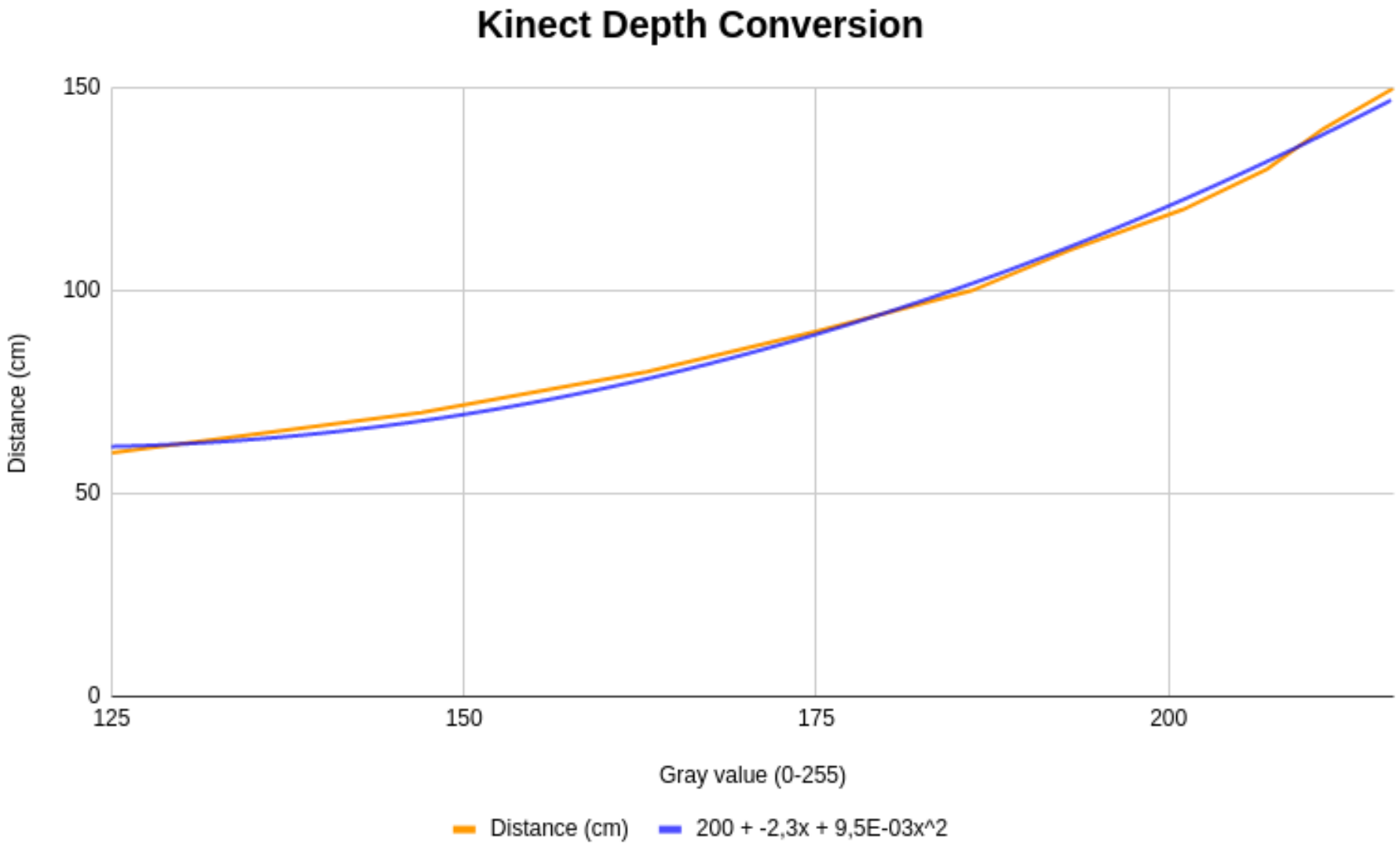

*Figure 2 - Microsoft Kinect XBOX 360 depth conversion from gray to distance*

A comparative analysis of the Intel® RealSense™ D455 and the Microsoft Kinect Xbox 360 cameras was conducted to assess their suitability for depth mapping in the proposed teleoperation system. The RealSense D455 offers higher depth accuracy, with a typical error of less than 2% at a range of up to 6 metres, along with improved depth resolution, which is essential for capturing fine details in human pose estimation. In contrast, the Kinect Xbox 360, while more cost-effective and easier to integrate due to its broad developer community, has lower depth resolution and reduced accuracy, particularly at greater distances, which can result in less reliable pose estimation. Furthermore, the RealSense D455 benefits from built-in support for the RGB-D pipeline and advanced SDK features, simplifying the integration with the Mediapipe framework. Given these considerations, the RealSense D455 was selected for its superior depth accuracy and resolution, ensuring precise human pose detection and enhancing the overall performance of the teleoperation system.

After comparing both cameras, the Intel® RealSense™ D455 demonstrated superior performance due to its higher accuracy range—an essential factor for precise depth measurements. Furthermore, the Intel camera provides higher image and depth resolution, making it a more desirable option than the Microsoft Kinect XBOX 360.

### 3.2 Body Pose Estimation via Image Processing

To accurately calculate all joint movements and determine the values for each degree of freedom, a Python algorithm was developed. MediaPipe was used to perform pose estimation on the colour frame, identifying the key points in 2D space. To extend this to 3D, the RealSense Python library's *get_distance()* function was employed, taking a specific pixel position as input and returning its depth value. This allowed the construction of a 3D model of the human body containing all key points.

With this 3D model, joint angles could be computed with greater precision. The chosen method for calculating these angles was the scalar (dot) product, represented by the following formula:

$$\theta = \cos^{-1}\left[\frac{A \cdot B}{\|A\|\|B\|}\right] \tag{1}$$

A and B are two vectors that share a common origin, from which the angle θ between them is measured. The scalar (dot) product of the two vectors is represented as A·B, with the norm of each vector denoted by ||A|| and ||B||, respectively.

The inclusion of depth information can also introduce a problem known as joint overlap, which can lead to inaccurate estimations. To address this, a rotatable array system can be implemented to store joint values. By rotating the array and averaging the stored values, any overlap can be accounted for, enabling more accurate joint depth estimation

While incorporating a third dimension significantly improves the precision and accuracy of calculating joint angles, there are cases where a two-dimensional approach may suffice. One such example is calculating the Shoulder Pitch. Using the 3D coordinates of the key points, these can be decomposed into distinct planes for analysis: the XY plane (frontal view), YZ plane (lateral view), and XZ plane (top view), creating orthogonal projections of the 3D structure, as illustrated in Figure 3. By focusing solely on the relevant planes for each analysis, the calculation process can be simplified and made more efficient. The colour scale in the images represents the depth map captured by the RGB-D camera, where each pixel's colour encodes the distance from the camera to the first object detected in the scene.

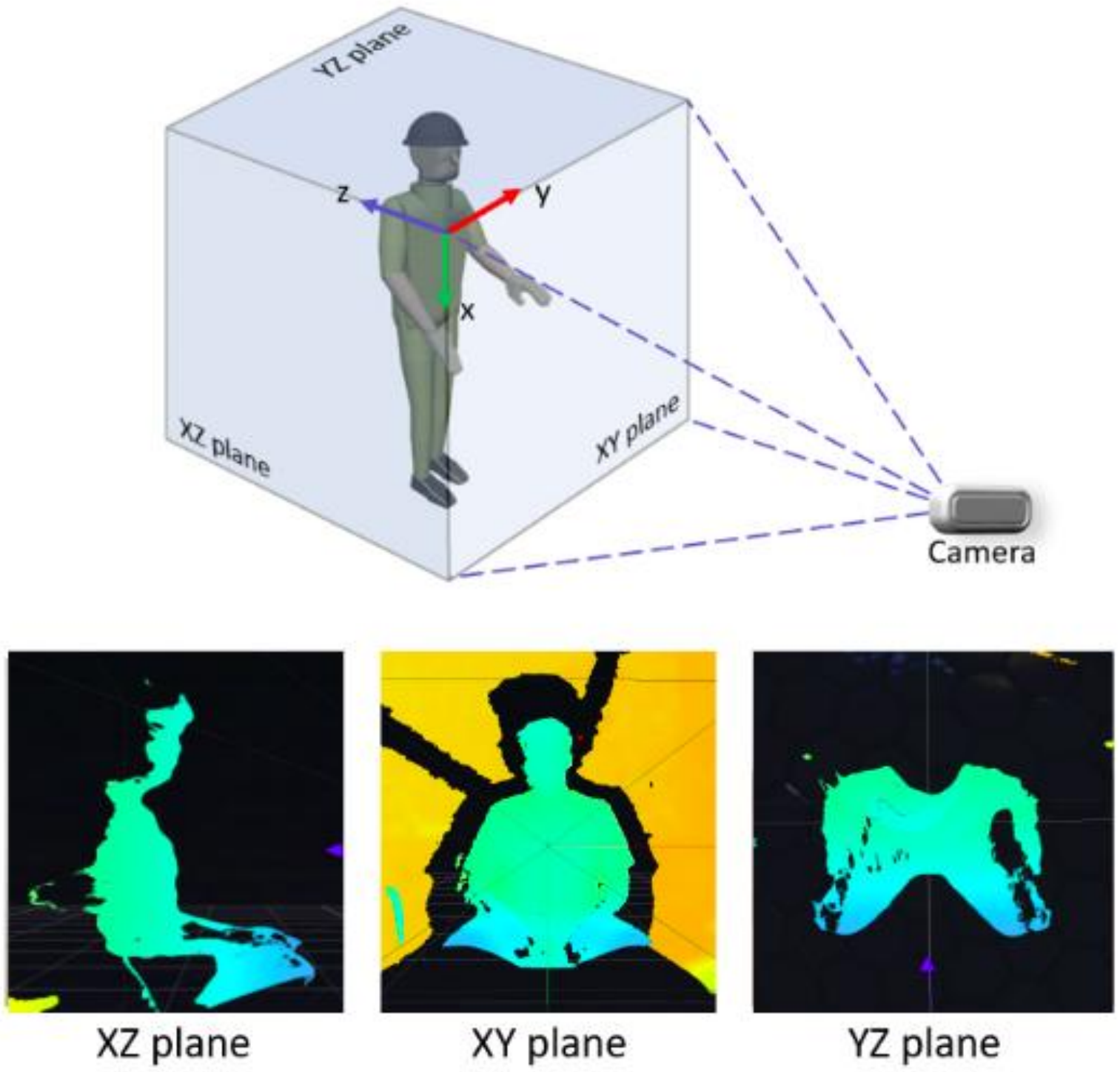

*Figure 3 - Orthogonal projection of the 3D view*

### 3.2.1 Shoulder Pitch Calculation

Shoulder Pitch refers to the shoulder's forward-backward movement along the Y-axis. To calculate this degree of freedom, three reference points are considered for the scalar product calculation: the hip, the shoulder, and the elbow. As it involves forward-backward motion, only the x and z coordinates of these points are used, which correspond to the lateral plane, as illustrated in Figure 4.

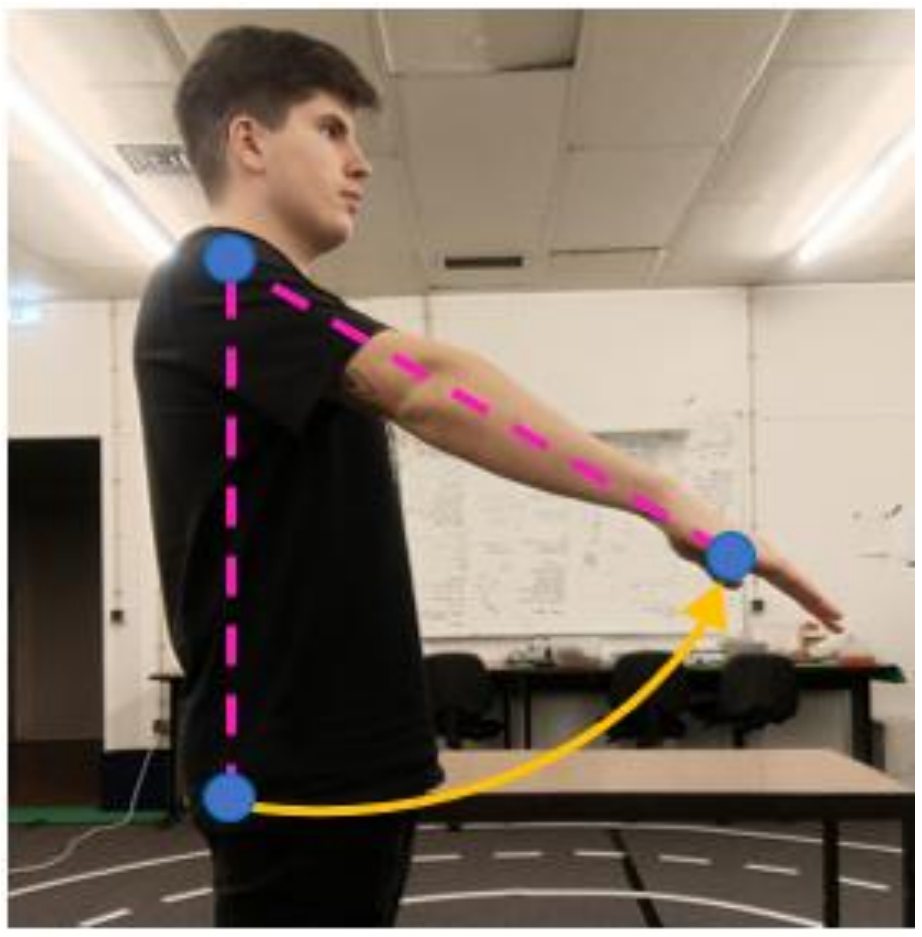

*Figure 4 - Representation of Shoulder Pitch calculation*

### 3.2.2 Shoulder Roll Calculation

Like Shoulder Pitch, Shoulder Roll is calculated using the same three reference points: the hip, the shoulder, and the elbow. However, Shoulder Roll describes a lateral movement; therefore, only the x and y coordinates are used, which correspond to the frontal plane, as shown in Figure 5.

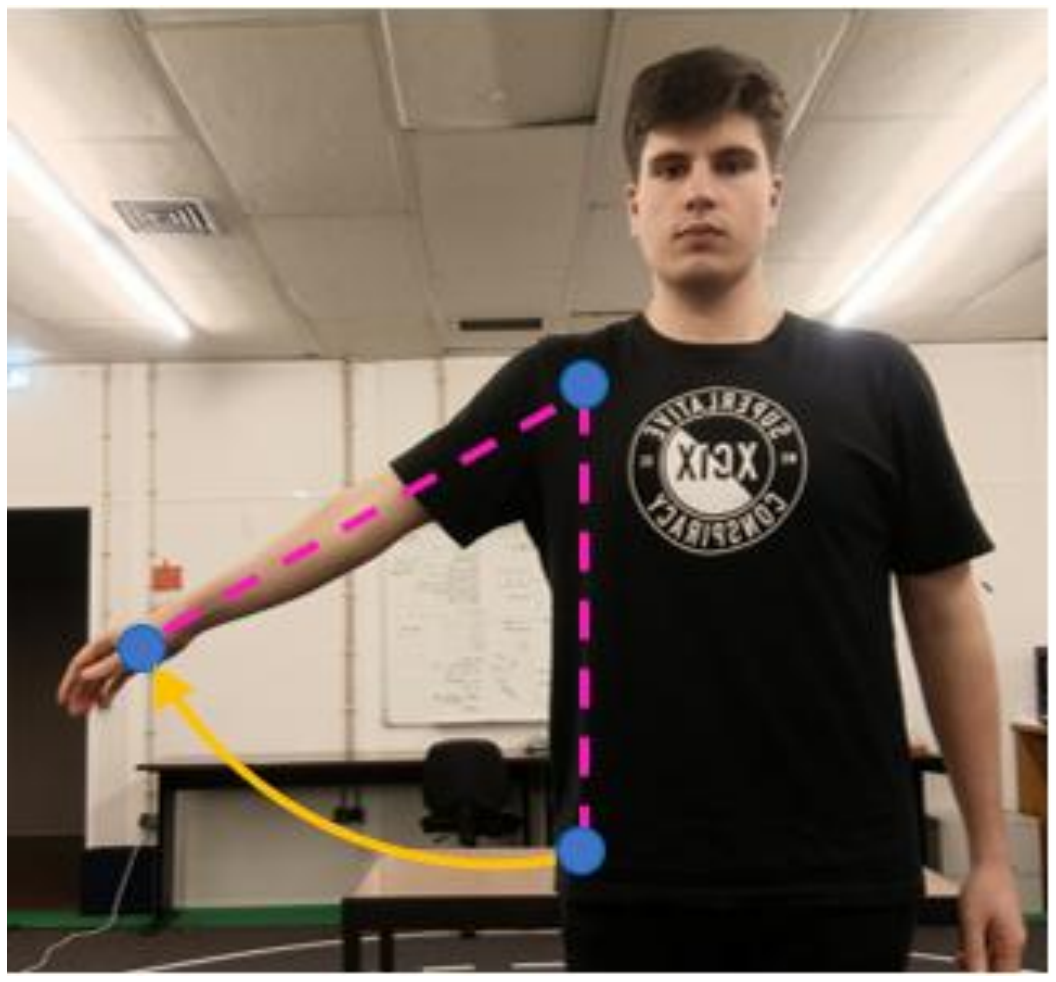

*Figure 5 - Representation of Shoulder Roll calculation*

One limitation of the Shoulder Roll calculation occurs when the elbow is positioned in front of the shoulder. In this situation, the angle may be miscalculated, resulting in an erroneous Shoulder Roll value. To address this, a new method was developed that incorporates the depth difference between the elbow and the shoulder, allowing the algorithm to adjust the Shoulder Roll measurement accordingly.

$$ShoulderRoll = ShoulderRoll \cdot \left(1 - \frac{1}{20} \cdot (ShoulderDepth - ElbowDepth)\right) \quad (2)$$

This is achieved by multiplying the Shoulder Roll value by a corrective factor (between 0 and 1) based on the depth difference.

#### 3.2.3 Elbow Pitch Calculation

Elbow Roll is the degree of freedom responsible for the open-close motion of the elbow. As with the previous degrees of freedom, the angle is calculated using the scalar product of three reference points: the shoulder, elbow, and wrist, as depicted in Figure 6.

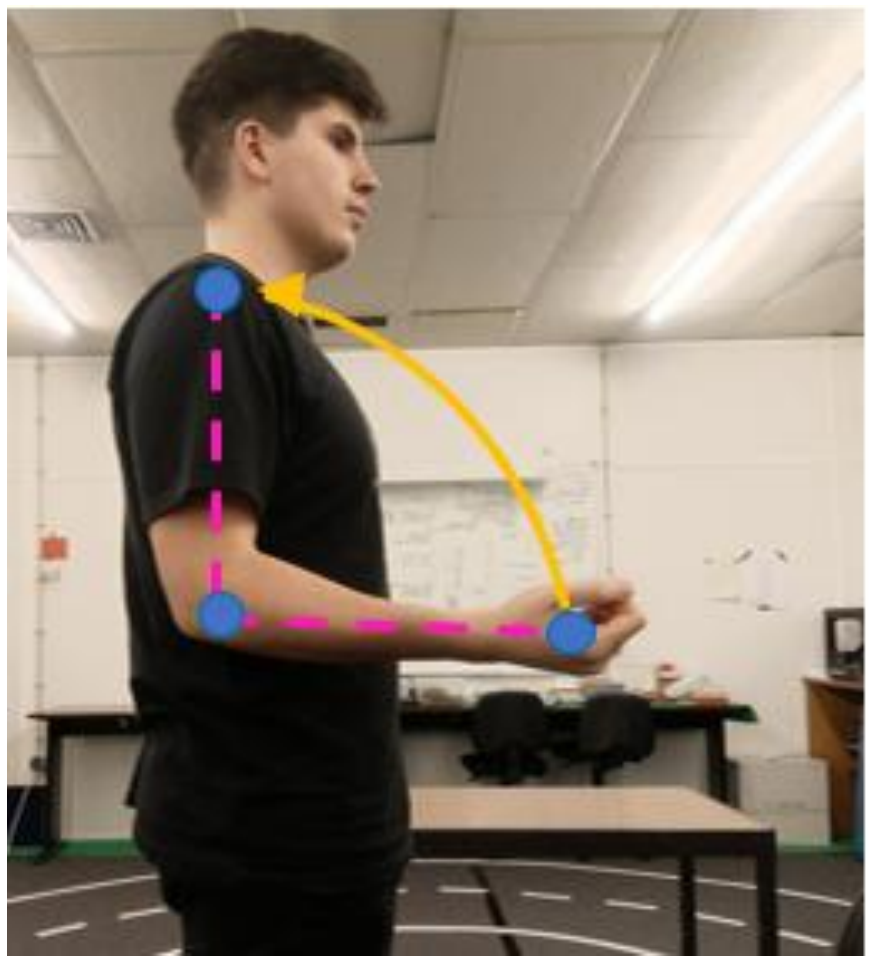

*Figure 6 - Representation of Elbow Roll calculation*

#### 3.2.4 Elbow Yaw Calculation

Elbow Yaw corresponds to the rotational motion of the elbow around the upper arm's axis. Calculating this motion can be challenging, as precise estimates from image processing alone are often insufficient. For this calculation, three points are used: the wrist, the elbow, and an auxiliary point that shares the x-coordinate of the elbow and the z-coordinate of the wrist, as shown in Figure 7.

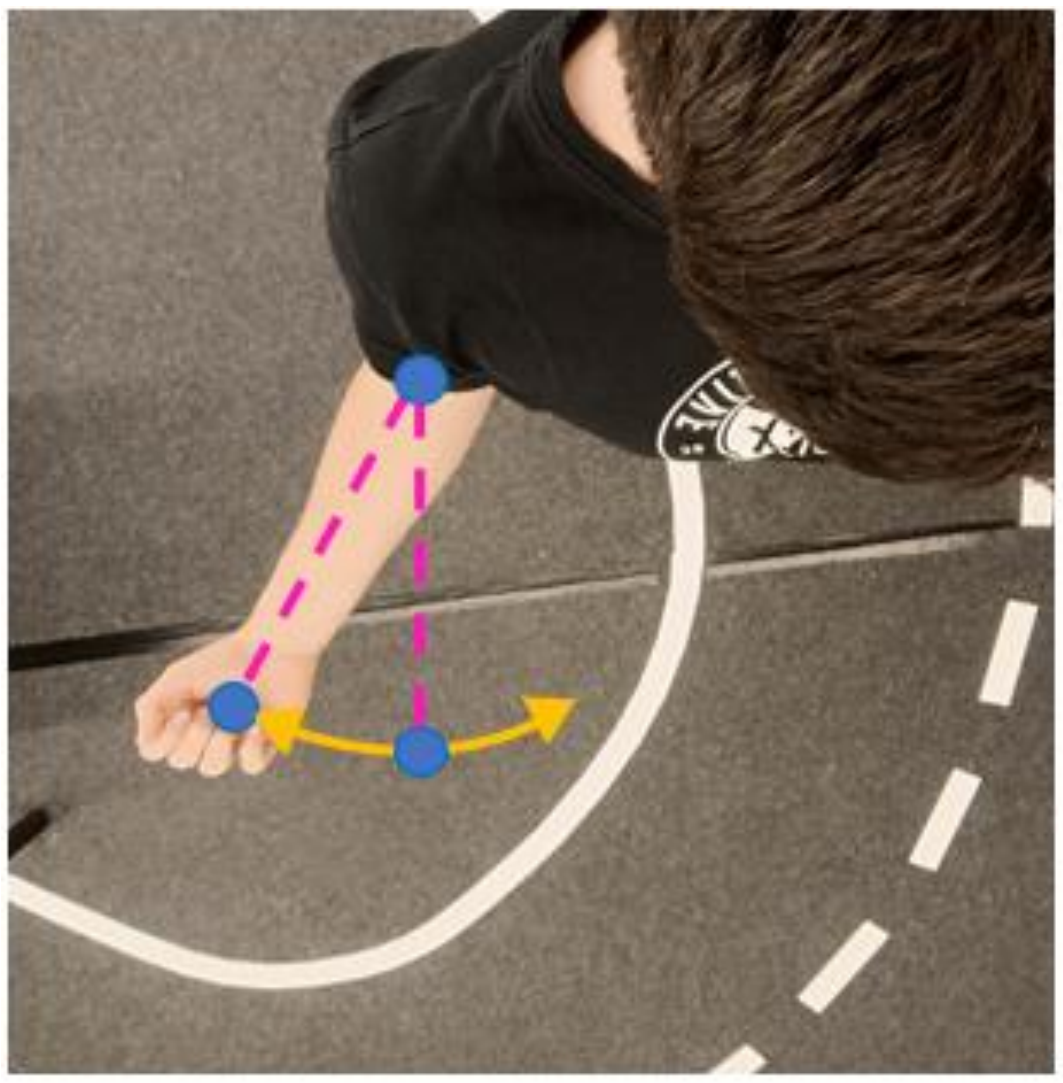

*Figure 7 - Representation of Elbow Yaw calculation*

A detailed analysis was conducted to explore all possible torsion scenarios, which can be categorized as either outward (positive) or inward (negative) rotation. The x and y coordinates of the shoulder were considered, resulting in four possible cases for locating the elbow's x and y coordinates relative to the shoulder. Furthermore, each of these cases was subdivided into four possibilities for the wrist's x and y coordinates relative to the elbow, yielding a total of 16 scenarios.

Figure 8 illustrates the possible Elbow Yaw configurations: subfigure (a) shows the right arm, while subfigure (b) shows the left arm. Red crosses mark physically impossible positions, while yellow crosses represent physically possible positions that cannot be achieved by the robot due to its limited joint range of motion.

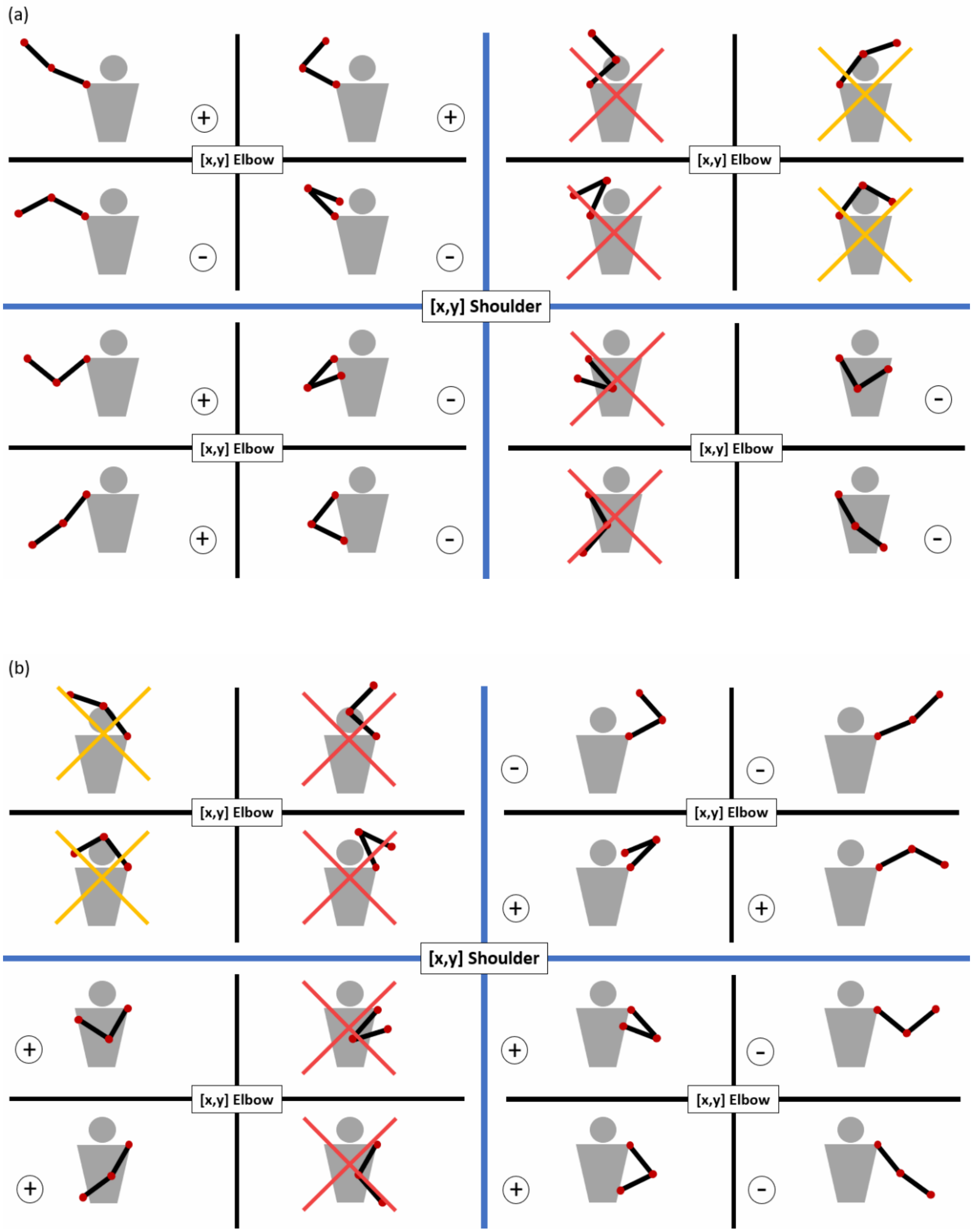

*Figure 8 - Right arm (a) and left arm (b) elbow yaw configurations*

#### 3.2.5 Wrist Yaw Calculation

Wrist Yaw is the degree of freedom that allows wrist rotation. Accurately determining this rotation using only image processing can be difficult. To address this challenge, a method was implemented that involves calculating the slope of two lines, both of which share the wrist as a common point. The first line connects the wrist to the metacarpophalangeal joint (MCP) of the index finger, while the second line connects the wrist to the MCP of the Pinky finger. The slope of each line is calculated using the following mathematical expression:

$$m = \frac{y_2 - y_1}{x_2 - x_1} \tag{3}$$

Where $m$ represents the slope of the line formed by the points $(x_1, y_1)$ and $(x_2, y_2)$.

After computing the slopes of both lines, the angle between them can be calculated using the arctangent of the difference in their slopes.

$$\theta = tan^{-1}\left[\frac{m_1 - m_2}{1 + m_1.m_2}\right] \tag{4}$$

Where $\theta$ represents the angle between the two slopes $m_1$ and $m_2$. Using both equations previously presented, the angle of the wrist can be determined, as illustrated in Figure 9.

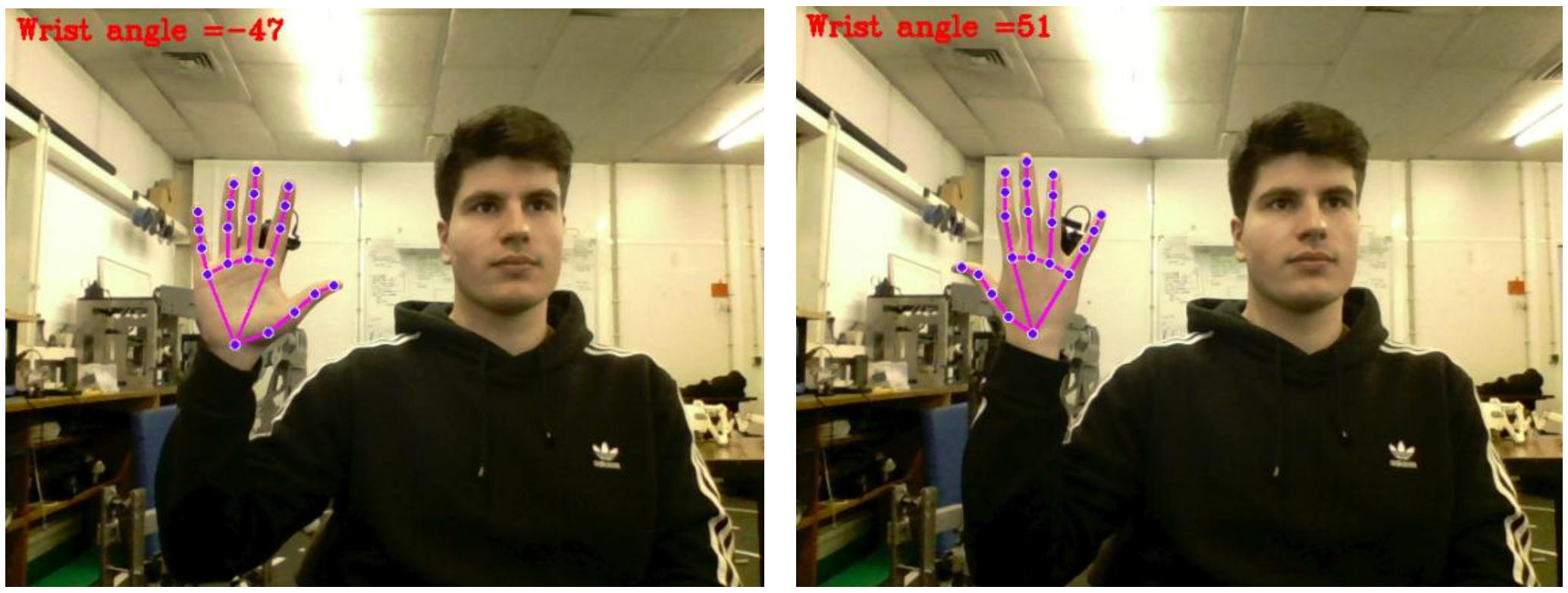

*Figure 9 - Wrist Yaw calculation*

Although this method can be effective in certain situations, it has limitations when the arm is oriented towards the camera—meaning the palm is not visible—which can complicate accurate wrist rotation estimation. Figure 10 shows three cases where the arm is in different positions.

In all cases, there is no actual wrist rotation; however, the slope values calculated for each case vary, demonstrating that this method is not consistently reliable for wrist rotation estimation.

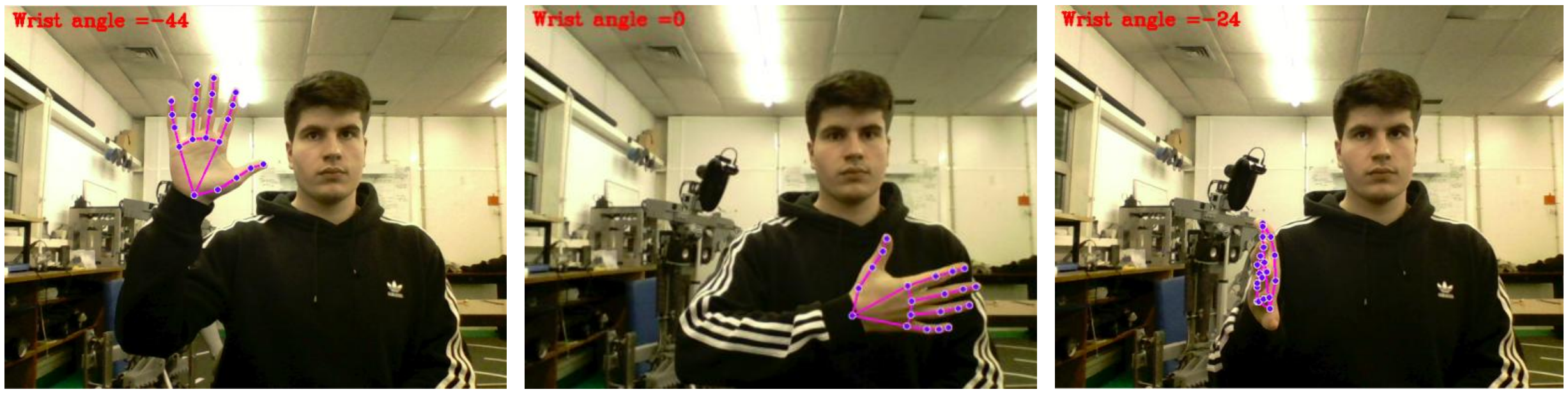

*Figure 10 - Failing cases for Wrist Yaw calculation*

An effective solution to this challenge is to utilize Inertial Measurement Units (IMUs). IMUs are equipped with advanced gyroscopes that can accurately determine wrist rotation with high precision.

### 3.2.6 Hand Opening and Closing

One of the key advantages of teleoperation is the ability to manipulate objects. This control includes two distinct modes: opening and closing. Individual finger control is not implemented in this system. To detect the opening and closing of the hand, the wrist and the middle fingertip are used as reference points. Using a 3D vector-based approach and the Euclidean distance formula, the distance between these two points is measured.

$$dist = \sqrt{(W_x - M_tip_x)^2 + (W_y - M_tip_y)^2 + (W_z - M_tip_z)^2} \quad (5)$$

With *W* being the wrist key point and *M_tip* being the middle fingertip key point.

## 4. Virtual Reality

Telepresence technology empowers individuals to remotely control and communicate with robotic devices located in distant environments. This technology creates an immersive experience, especially when combined with virtual reality (VR) equipment, enabling users to feel as though they are physically present at the remote site.

The PX2.0 virtual reality headset offers a highly satisfactory VR experience at an affordable price. Equipped with adjustable lenses, it allows users to transform two separate images displayed on a smartphone into an immersive VR environment by simply placing the phone into the headset's housing. These images are processed in OpenCV and streamed to a web framework, making them accessible on the phone.

However, one challenge emerged during the implementation of the VR headset: when the headset was worn by the operator, the face could not be detected, rendering head pose estimation ineffective. To overcome this limitation, a surface with a distinguishable colour was placed in front of the headset. An HSV mask filter was then applied to detect only the target colour. This mask was processed to extract the contours and compute the centroid, which was approximately aligned with the nose. Figure 11 shows the process of mask extraction and contour detection, with the centre of the contour represented by a white dot.

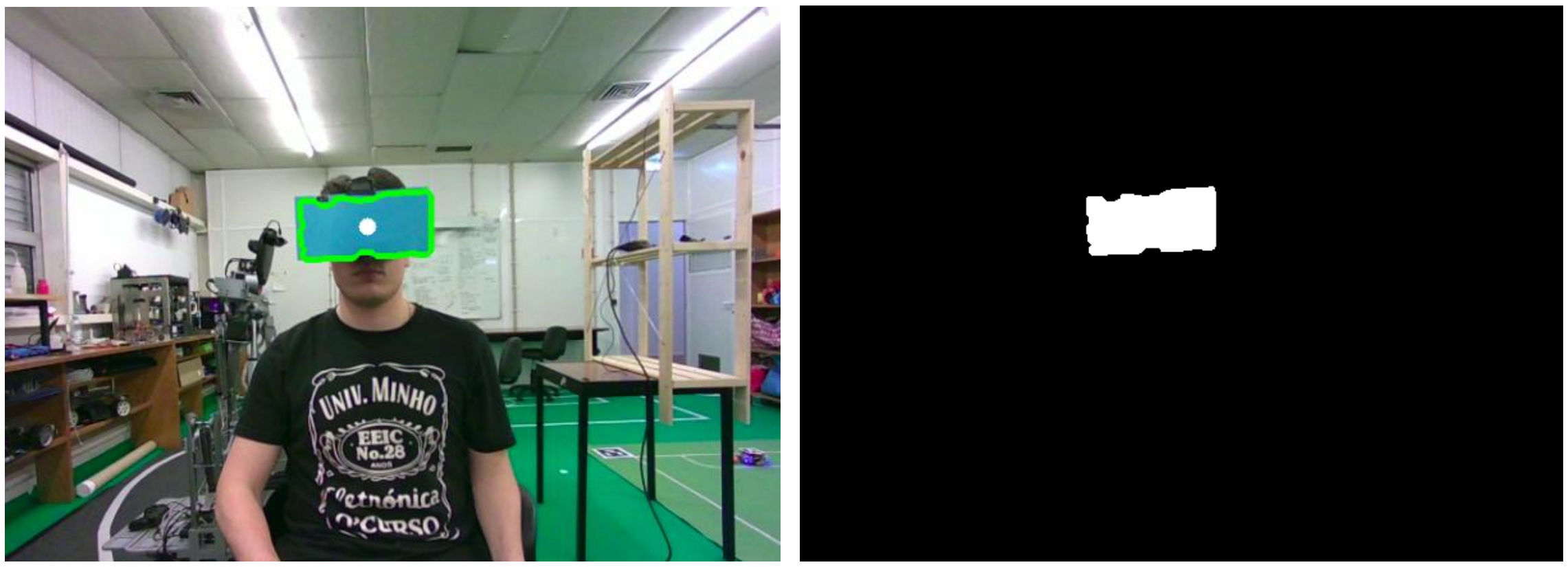

*Figure 11 - Blue colour contour with centre point (left) and respective blue colour mask (right)*

Once the centroid was determined, depth information was retrieved to obtain the three-dimensional coordinates. For calculating the head pitch, two additional points were created: one at the centre of the shoulders and another at half the shoulder width, maintaining the same depth and positioned perpendicular to the first point. These three points — the mask centre and the two additional points — were then used to calculate the head pitch angle. Figures 12(a) and 12(b) illustrate the placement of these points and the head pitch angle calculation.

For head yaw estimation, a similar approach to the elbow yaw calculation was employed. The first point was the one above the shoulder line, while the second point shared the same x and y coordinates but had a lower depth value. The angle between these points, along with the mask centre, was used to determine the head yaw angle in the top-down (XZ) plane. Figure 12(c) shows the arrangement of these three points and the calculation of the head yaw angle.

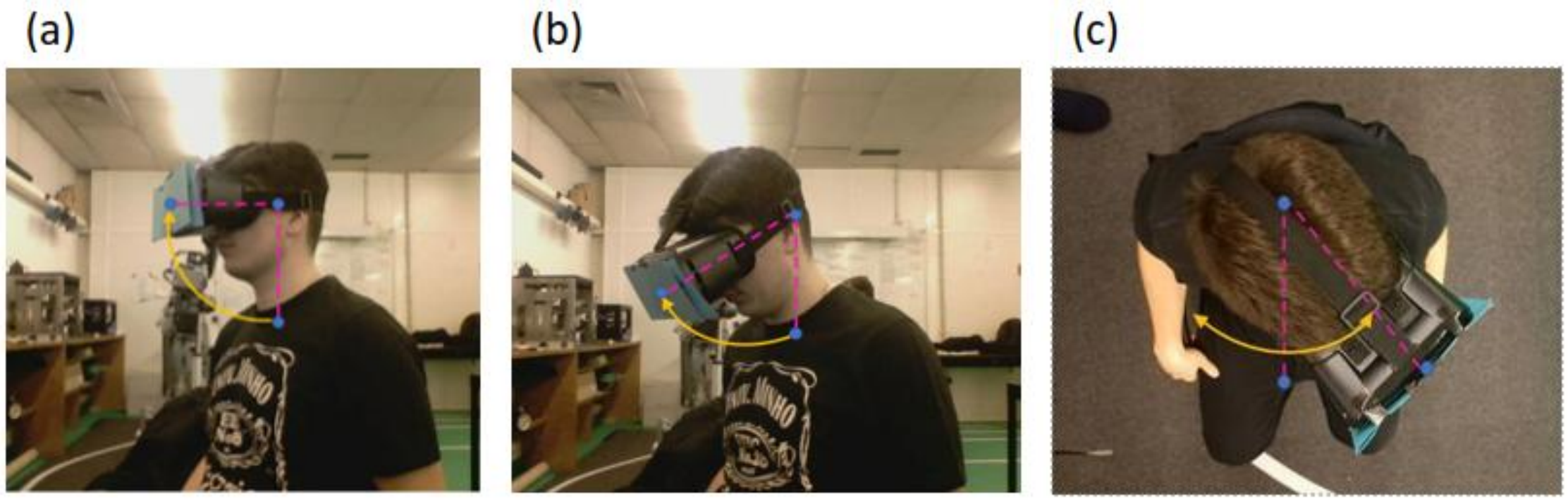

*Figure 12 - Head pitch (a, b) and head yaw (c) calculation*

## 5. Real World Tests

Teleoperating a robot equipped with an anthropomorphic arm presents both unique challenges and opportunities. While this configuration enables the robot to execute tasks that demand high dexterity and precision, it also requires a reliable and responsive communication system to ensure that the operator's commands are effectively transmitted and executed.

### 5.1 Controlling the Anthropomorphic Arm

The initial test aimed to evaluate the functionality of CHARMIE's robotic arm [13]. For this test, modifications were made to the existing code, particularly to the communication protocols with the Arduino Mega. As shown in Figure 13, the test setup established communication between the various devices involved in the control of the arm.

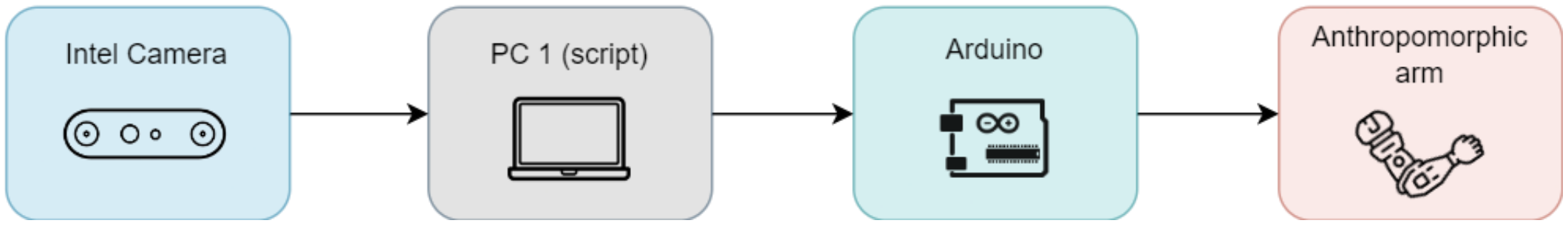

*Figure 13 - Communication between devices during the anthropomorphic arm control test*

Given that the values sent to the servos vary continuously, applying these values directly could result in unstable operation. To address this, a filtering approach was implemented to smooth the data. A cyclic array updates the values at each iteration, and an averaging filter is applied to ensure that the servo receives stable angle values.

Another critical consideration when sending values to the servos is the limitation of their range of motion. The calculated angles may exceed the servos' physical limits; hence, an interpolation function is used to map the calculated angle into the servo's operational range. This function requires five parameters: the calculated angle, the minimum and maximum values of the calculated angle, and the minimum and maximum values allowed by the servo.

$$Servo_value = max\begin{cases} min\begin{cases} out_{max} \\ \frac{(x-in_{min})\cdot(out_{max}-out_{min})}{in_{max}-in_{min}} + out_{min} \end{cases} \\ out_{min} \end{cases} \quad (6)$$

Where *Servo_value* is the mapped servo value, *x* represents the calculated value, $in_{min}$ and $in_{max}$ denote the minimum and maximum values for the calculated angle, and $out_{min}$ and $out_{max}$ correspond to the minimum and maximum values permitted for the servo's range of motion.

The objective of this test was to evaluate the effectiveness of CHARMIE's object manipulation capabilities using Intel RealSense technology. The Intel RealSense camera was mounted atop CHARMIE's head to capture real-time data, while the operator positioned in front of the robot supervised the system. The results of this initial test are illustrated in Figure 14, demonstrating that CHARMIE was able to manipulate the object successfully, performing the task with both precision and accuracy.

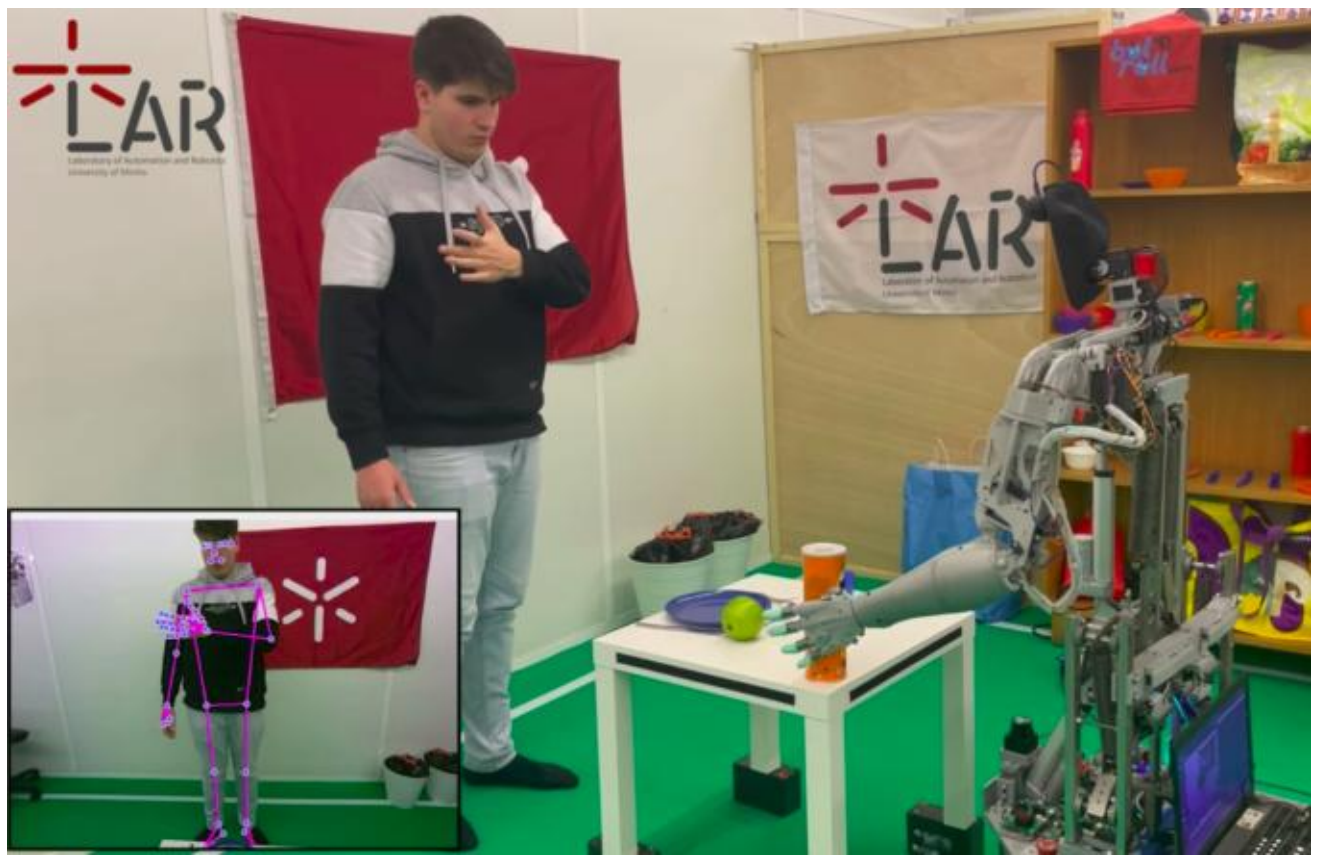

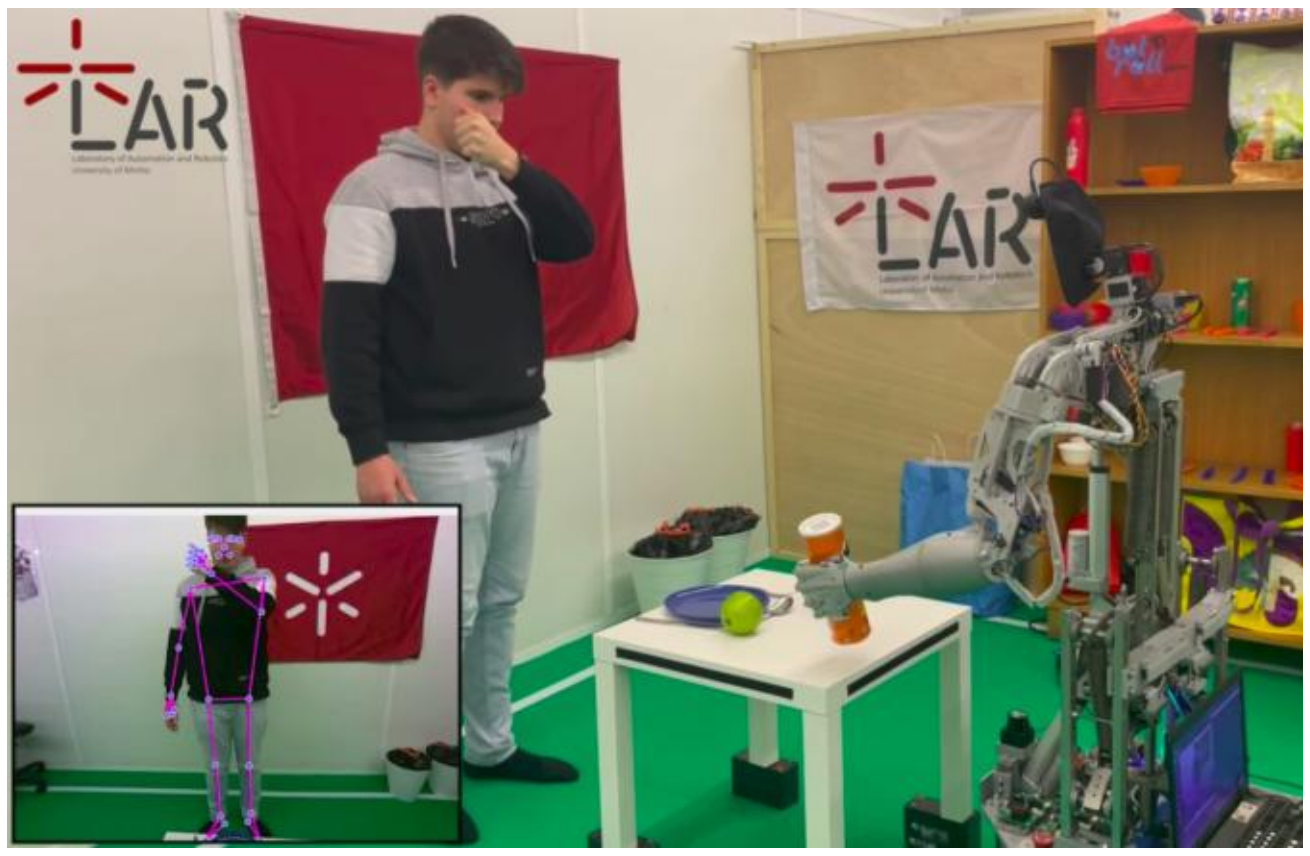

*Figure 14 - Object manipulation test [images from [15]]*

### 5.2 Controlling the robot

The final test involves the operator manipulating an object using virtual reality while simultaneously controlling CHARMIE's robotic arm. The operator gains real-time access to the robot's vision, controlling CHARMIE's neck movements by tilting their own head to explore the environment in 3D and perform precise manoeuvres. Additionally, the robot's arm is guided directly by the operator's movements, enabling fine and delicate manipulation. This test effectively evaluates the operator's skill and the system's responsiveness, creating an immersive and interactive experience.

Figure 15 presents a diagram illustrating the communication architecture among the various devices involved in this test. The operator is positioned in front of the Intel camera, which captures images and sends them to PC 1 for processing. The joint angle values are then transmitted from PC 1 to PC 2 via socket communication. PC 2 subsequently relays these values to CHARMIE's arm and neck actuators. Simultaneously, PC 2 receives images from a camera mounted on the robot and performs stereoscopic rendering for the virtual reality interface. These images are streamed to the operator's VR headset, enabling a fully immersive teleoperation experience.

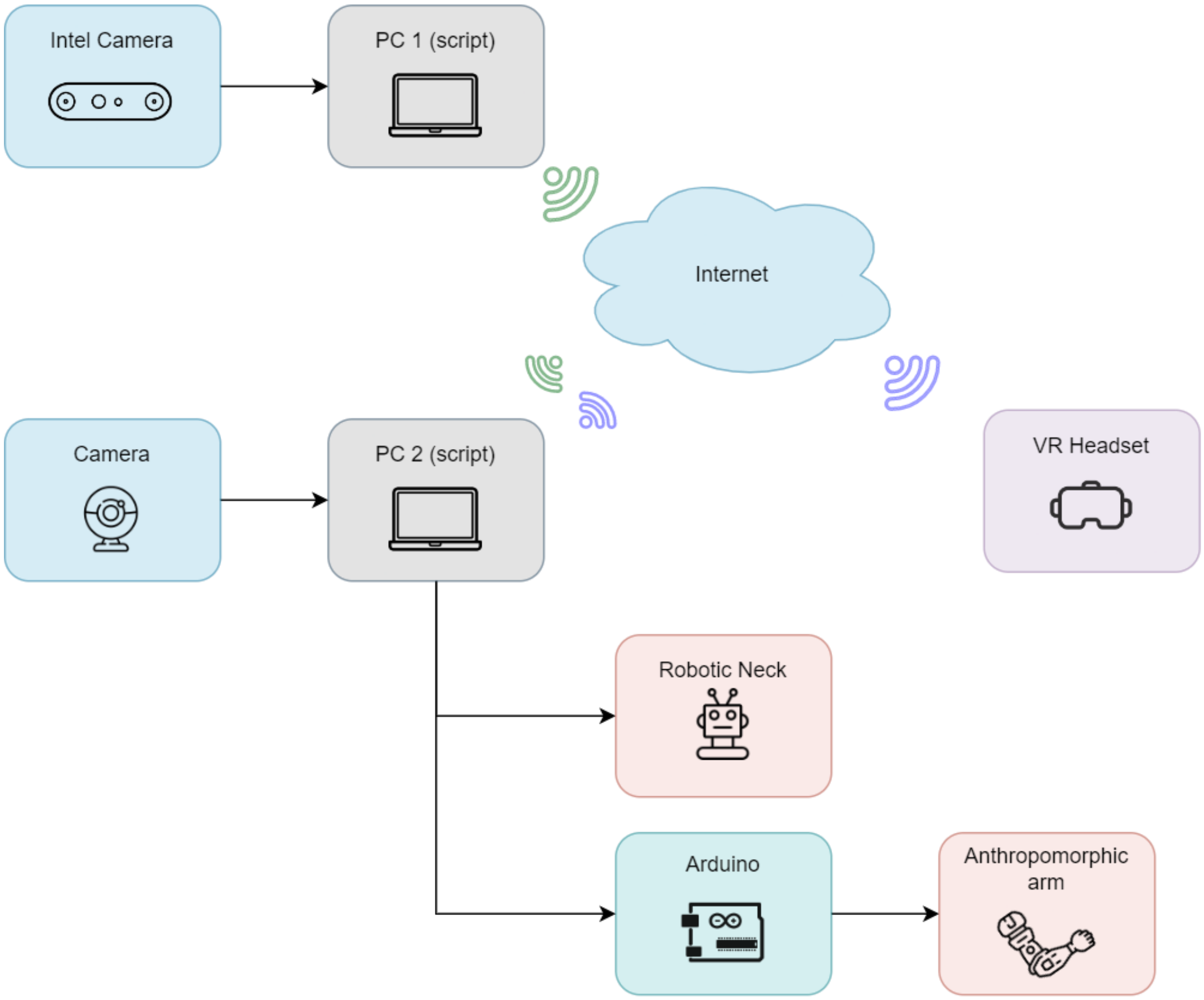

*Figure 15 - Communication between devices on robot control test*

The results of the test are illustrated in Figure 16, which presents three distinct action frames showcasing the robot's perspective and corresponding pose detection. The first frame (Figure 16 (a)) shows the arm approaching the can, while the second frame (Figure 16 (b)) depicts the arm in contact with the can. Finally, the third frame (Figure 16 (c)) demonstrates the successful grasping of the can. These frames highlight the effectiveness of the teleoperation system in achieving precise and reliable object manipulation.

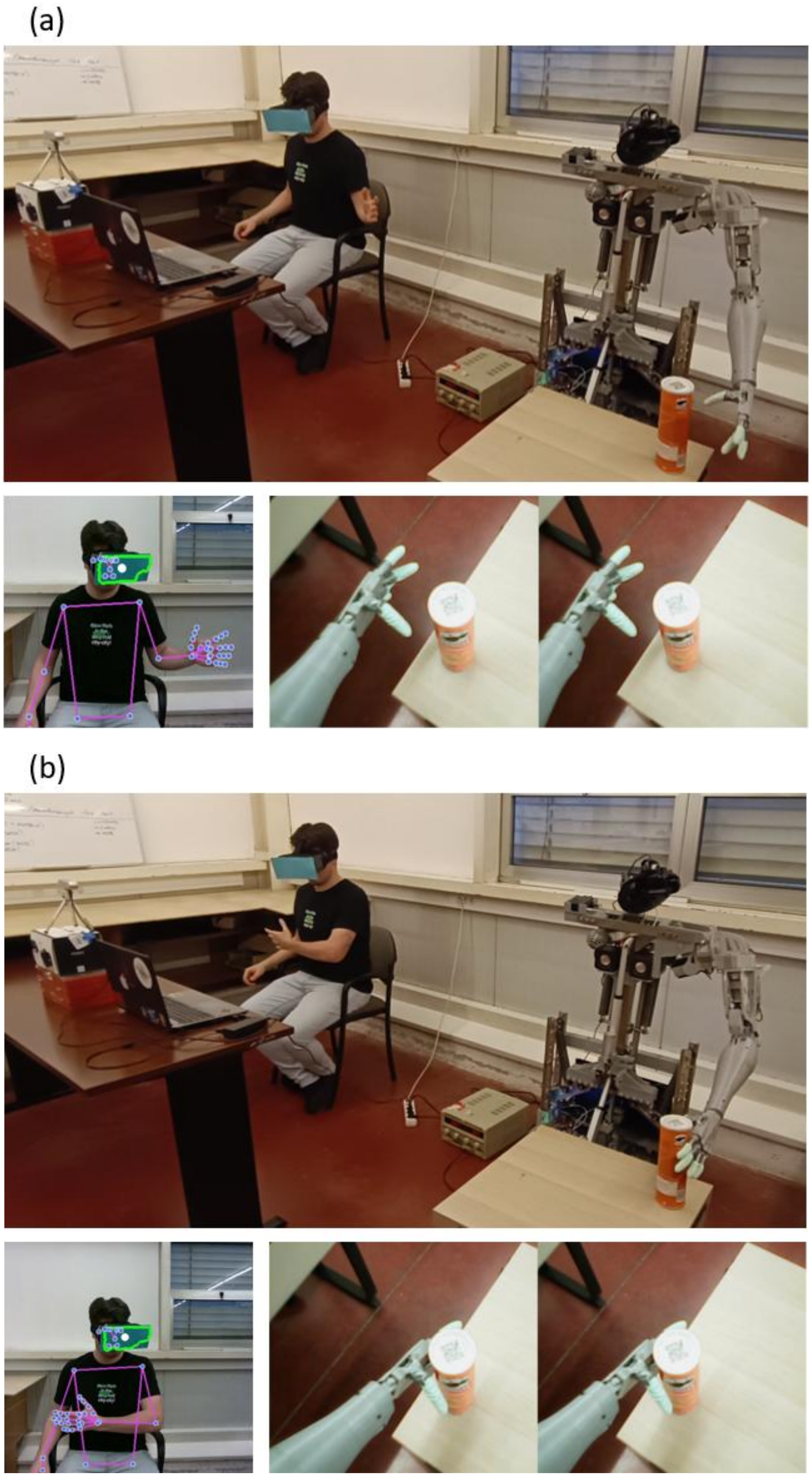
(a)
(b)

(c)

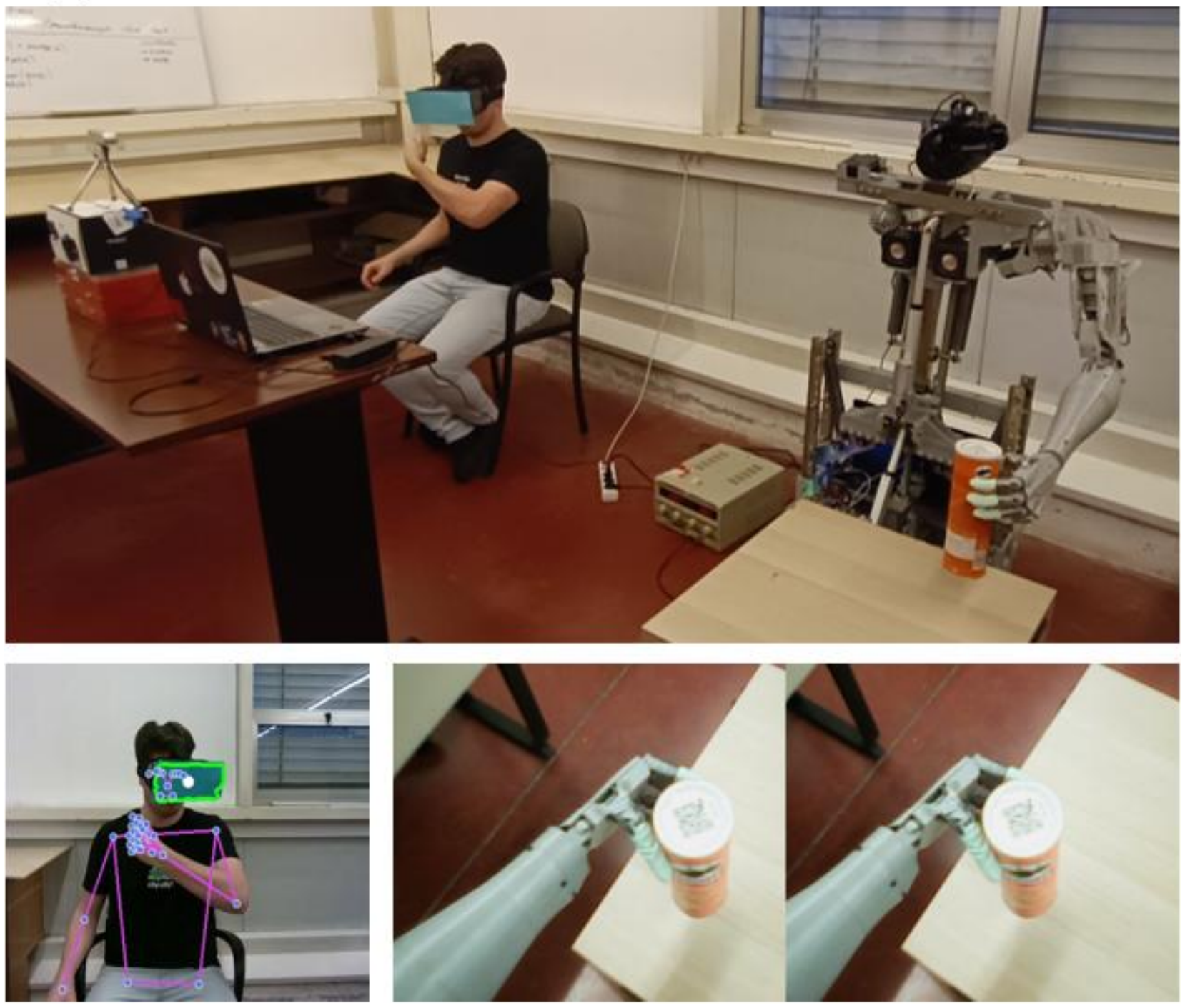

*Figure 16 - Object manipulation test through VR (Reaching the object (a,b) and picking up the object (c))*

## 6. Conclusions and Future Work

This work demonstrates the successful teleoperation of the CHARMIE robot using a robust pose estimation algorithm, enabling precise object manipulation and human–robot interaction. The experiments were conducted seamlessly, showcasing the system's capabilities in real-world scenarios. However, the results also highlight the limitations of relying solely on image processing, particularly in accurately estimating the Wrist Yaw joint rotation.

To further enhance the system's performance and improve human pose detection in virtual reality environments, several avenues for future research and development are identified:

1. Implementation of advanced rotational calculation techniques—specifically for Wrist Yaw and Elbow Yaw—by integrating Inertial Measurement Units (IMUs) to enhance rotational accuracy.
2. Utilization of gyroscope data from the user's smartphone, integrated with the VR headset, to achieve more precise head movement tracking and immersive control.

3. Development of a user-friendly interface that seamlessly displays the robot's outputs within the virtual environment, thereby enhancing the operator's experience and improving usability.

4. Integration of a large-scale tactile sensing system on the robot's forearm to provide real-time feedback on contact and interactions, enabling safer and more intuitive object manipulation [16].

These improvements will contribute to the development of a more effective and immersive teleoperation system, paving the way for more reliable and versatile service robots in diverse application domains.

Although the teleoperation system presented in this paper demonstrates the feasibility of controlling a service robot using human pose estimation, some limitations were identified—particularly in aligning the human arm's movement with that of the robot's arm. As indicated by the reviewer, Figure 16 highlights occasional discrepancies between the human operator's movements and the robot's corresponding actions. To address this, future work will focus on refining the mapping algorithm to improve consistency between human and robot movements. This includes incorporating advanced filtering techniques, dynamic adaptation to the robot's mechanical constraints, and learning-based approaches to fine-tune the mapping function in real time. Additionally, integrating feedback mechanisms from the robot's sensors could allow the system to correct errors on the fly, ensuring a more accurate replication of human movements.

## Acknowledgements

The authors would like to acknowledge the Laboratory of Automation and Robotics (University of Minho, Guimarães) for their support, equipment, and valuable staff assistance.

This work has been supported by FCT – Fundação para a Ciência e Tecnologia within the R&D Units Project Scope: UID/00319/2023.